\documentclass[11pt,a4paper]{article}
\usepackage{hyperref}
\usepackage[hyperref]{emnlp2020}
\usepackage{times}
\usepackage{latexsym}

\usepackage{graphicx}
\usepackage[normalem]{ulem}
\usepackage{amsmath}

\usepackage{makecell}
\usepackage{cleveref}
\usepackage{multirow}
\usepackage{ulem}

\crefformat{footnote}{#2\footnotemark[#1]#3}

\usepackage{microtype}

\aclfinalcopy 
\def\aclpaperid{118} 

\title{NeuSpell: A Neural Spelling Correction Toolkit}

\author{Sai Muralidhar Jayanthi, Danish Pruthi, Graham Neubig \\
Language Technologies Institute \\
Carnegie Mellon University \\
\texttt{\{sjayanth, ddanish, gneubig\}@cs.cmu.edu}
}

\date{}

\crefname{section}{§}{§§}
\Crefname{section}{§}{§§}

\begin{document}
\maketitle

\begin{abstract}
We introduce NeuSpell, an open-source toolkit
for 
spelling correction in 
English.
Our toolkit 
comprises ten 
different models,
and 
benchmarks them on
naturally occurring misspellings 
from multiple sources.
We find that many
systems 
do not 
adequately 
leverage the context around the 
misspelt token.
To remedy this, 
(i) 
we train neural models using
spelling errors in context, 
synthetically constructed by reverse engineering 
isolated misspellings;
and (ii)
use contextual representations.
By training on our synthetic
examples, 
correction rates 
improve 
by  
$9$\% (absolute)
compared to the case when models are trained on
randomly sampled character perturbations.
Using richer contextual representations
boosts the correction rate by another $3$\%.
Our toolkit enables
practitioners to use 
our proposed and existing 
spelling correction systems, 
both via 
a unified command line, 
as well as 
a web interface. 
Among many potential applications, we demonstrate 
the 
utility of our spell-checkers  
in combating adversarial misspellings.
The toolkit can be accessed at 
\href{neuspell.github.io}{neuspell.github.io}.\footnote{Code and pretrained models are available at:  \href{https://github.com/neuspell/neuspell}{https://github.com/neuspell/neuspell}}

\end{abstract}

\section{Introduction}
\label{sec:introduction}

Spelling mistakes constitute
the largest share of errors
in written 
text~\cite{10.1007/s10791-006-9002-8, flor-futagi-2012-using}.
Therefore,
spell checkers
are ubiquitous, forming  
an integral part of many
applications
including  
search engines, 
productivity and collaboration tools,
messaging platforms, etc.
However, many well 
performing spelling correction
systems
are developed by corporations, trained 
on massive proprietary user data. 
In contrast, many freely available 
off-the-shelf correctors 
such as Enchant~\cite{enchant}, 
GNU Aspell~\cite{aspell},
and JamSpell~\cite{jamspell},
do not effectively use the 
context of the misspelled word.
For instance, they fail to disambiguate
\uwave{thaught} to
\emph{taught}
or \emph{thought}
based on the context:
\begin{figure}
\centering
\includegraphics[width=0.5\textwidth]{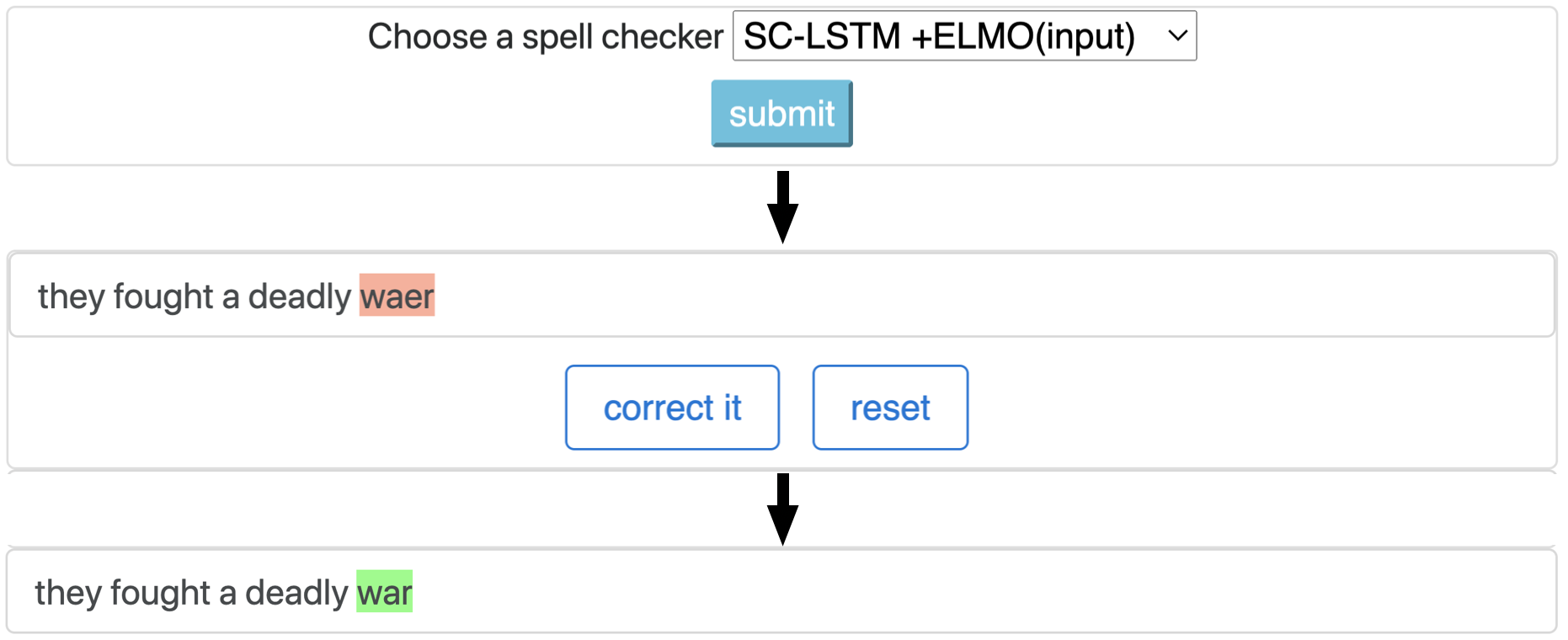}
\smallbreak
\includegraphics[width=0.5\textwidth]{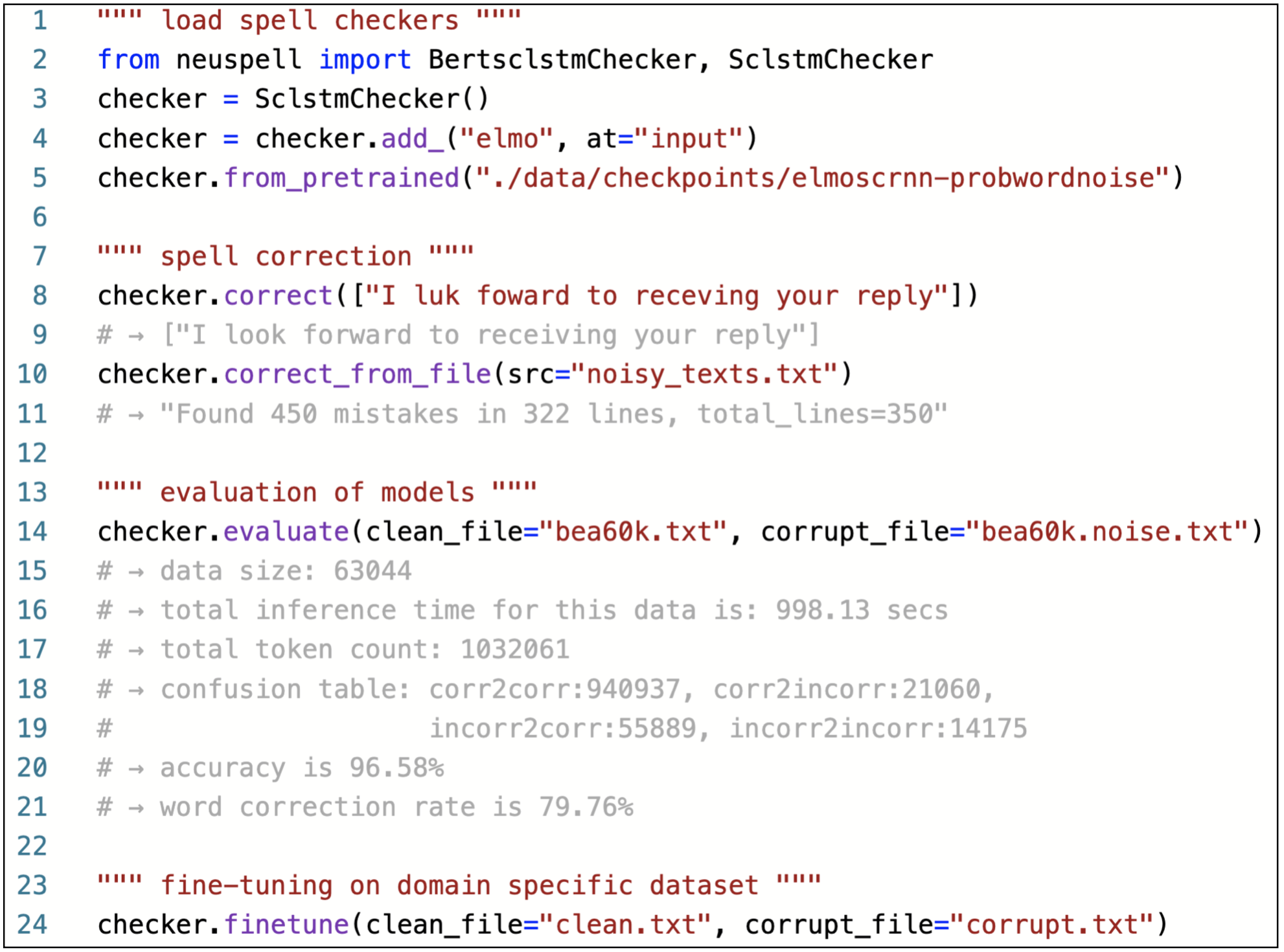}
\caption{\label{fig:ui_1} 
Our toolkit's web
and command line interface for spelling correction. 
}
\end{figure}
``Who \uwave{thaught} you calculus?" versus
``I never \uwave{thaught} I would be awarded the fellowship."

In this paper, we 
describe
our spelling correction toolkit, 
which comprises of several neural models
that accurately capture context
around the misspellings.
To train our neural spell correctors,
we first 
curate 
synthetic training data 
for spelling correction \emph{in context}, 
using several text noising strategies.
These strategies use a lookup 
table for word-level noising,
and
a context-based
character-level confusion dictionary 
for character-level noising.
To populate this lookup table and confusion matrix,
we harvest
isolated 
misspelling-correction 
pairs
from various publicly
available sources.

Further, we investigate 
effective ways to 
incorporate contextual information:
we experiment with contextual 
representations 
from pretrained models
such as
ELMo \citep{Peters_2018} 
and 
BERT \citep{devlin2018bert} 
and compare their 
efficacies with existing 
neural architectural choices
(\cref{ssec:exp_spell_correction}). 

Lastly, several recent studies
have 
shown that many 
state-of-the-art neural 
models 
developed for a 
variety of 
Natural Language Processing 
(NLP)
tasks easily break in the 
presence of 
natural or synthetic spelling 
errors~\cite{belinkov2017synthetic, ebrahimi2017hotflip, Pruthi_2019}.
We determine the usefulness 
of our toolkit
as a countermeasure against 
character-level adversarial attacks 
(\cref{ssec:exp_defense}).
We find that
our models 
are better defenses to adversarial attacks
than 
previously proposed spell checkers.
We believe that our toolkit
would 
encourage practitioners
to incorporate spelling correction systems 
in other
NLP applications.

\begin{table}[h]
\resizebox{0.48\textwidth}{!}{%
\begin{tabular}{lcc}
\hline
\multicolumn{1}{c}{Model} & 
\multicolumn{1}{c}{\makecell[c]{Correction\\Rates}} & \multicolumn{1}{c}{\makecell[c]{Time per sentence\\ (milliseconds)}} \\ 
\hline
\textsc{Aspell} \cite{aspell}
& 48.7 & 7.3$^{*}$ \\ 
\textsc{JamSpell} \cite{jamspell}
& 68.9 & \textbf{2.6}$^{*}$\\ 
\hline
\textsc{char-cnn-lstm} \cite{kim2015characteraware} 
& 75.8 & 4.2 \\ 
\textsc{sc-lstm} \cite{sakaguchi2016robsut} 
&  76.7 & \textbf{2.8} \\ 
\textsc{char-lstm-lstm} \cite{li2018spelling} 
& 77.3  & 6.4 \\ 
\textsc{Bert} \cite{devlin2018bert}
& 79.1 & 7.1 \\ 
\hline
\textsc{sc-lstm} &  &  \\
\hspace{3mm}+\textsc{Elmo} (input) & \textbf{79.8} & 15.8 \\ 
\hspace{3mm}+\textsc{Elmo} (output) & 78.5 & 16.3 \\ 
\hspace{3mm}+\textsc{Bert} (input) & 77.0 & 6.7 \\ 
\hspace{3mm}+\textsc{Bert} (output) & 76.0 & 7.2\\ 
\hline
\end{tabular}}
\caption{Performance of different correctors in the NeuSpell toolkit on the \textsc{BEA-60K} dataset with real-world spelling mistakes. $^*$ indicates evaluation on a CPU (for others we use a GeForce RTX 2080 Ti GPU).}
\label{tab:table-0-unnamed}
\end{table}

\section{Models in NeuSpell}
\label{sec:models}

Our toolkit offers 
ten different spelling correction models, 
which include:
(i) two off-the-shelf non-neural models,
(ii) four published neural models for spelling correction, 
(iii) four of our extensions. 
The details of first six  
systems are following:
\begin{itemize}
    \item GNU Aspell~\cite{aspell}:
    It uses a combination of 
    metaphone phonetic
    algorithm,\footnote{\label{metaphone_algo}\href{http://aspell.net/metaphone/}{http://aspell.net/metaphone/}}
    Ispell's
    near miss strategy,\footnote{\href{https://en.wikipedia.org/wiki/Ispell}{https://en.wikipedia.org/wiki/Ispell}} and a
    weighted edit distance metric to
    score candidate
    words.
    \item JamSpell~\cite{jamspell}:
    It uses a variant of the SymSpell algorithm,\footnote{\href{https://github.com/wolfgarbe/SymSpell}{https://github.com/wolfgarbe/SymSpell}}
    and a 3-gram language model
    to prune  
    word-level corrections.
    \item \textsc{sc-lstm}~\cite{sakaguchi2016robsut}:
    It corrects misspelt words 
    using semi-character 
    representations,
    fed through a
    bi-LSTM network.
    The semi-character representations
    are 
    a concatenation of 
    one-hot embeddings for the 
    (i) first, (ii)  last,  
    and (iii) bag of internal characters.
    \item \textsc{char-lstm-lstm}~\cite{li2018spelling}:
    The model builds word 
    representations 
    by passing its individual characters
    to a bi-LSTM. These representations
    are further fed to another bi-LSTM
    trained to predict the correction.
    \item \textsc{char-cnn-lstm}~\cite{kim2015characteraware}:
    Similar to the previous model, this model 
    builds word-level representations
    from individual characters using
    a convolutional network.
    \item \textsc{Bert}~\cite{devlin2018bert}:
    The model uses a pre-trained transformer network.
    We average the sub-word representations to obtain the word representations, which are further fed to a classifier to predict its correction.
\end{itemize}

To better capture the context 
around a misspelt token, 
we extend the \textsc{sc-lstm} model
by augmenting it with 
deep contextual representations
from pre-trained ELMo and BERT.
Since the 
best point to integrate 
such embeddings might
vary by task~\cite{Peters_2018},
we 
append
them
either to semi-character embeddings 
before feeding them to the biLSTM
or
to the biLSTM’s output.
Currently, our toolkit provides four such
trained models:
ELMo/BERT tied at input/output 
with a semi-character based bi-LSTM model.

\paragraph{\textbf{Implementation Details}}
Neural models in NeuSpell
are trained
by posing 
spelling correction
as a sequence labeling task, 
where a correct word is 
marked as itself and
a misspelt token is 
labeled as its
correction.
Out-of-vocabulary labels 
are marked as \texttt{UNK}.
For each word in the input 
text sequence, models 
are trained to output a 
probability distribution 
over a finite vocabulary 
using a softmax layer.

We set the hidden size of 
the bi-LSTM network in all models
to 512 and use \{50,100,100,100\} 
sized convolution filters with lengths
\{2,3,4,5\} respectively in CNNs. 
We use a dropout of 0.4 on the bi-LSTM's outputs
and train the models using cross-entropy loss.
We use the 
BertAdam\footnote{\href{https://github.com/cedrickchee/pytorch-pretrained-BERT/blob/master/pytorch_pretrained_bert/optimization.py}{github.com/cedrickchee/pytorch-pretrained-BERT}}
optimizer for models with a BERT component
and the Adam \citep{kingma2014adam} optimizer for the remainder. 
These optimizers are used with default parameter settings.
We use a batch size of 32 examples, and train with a patience of 3 epochs.

During inference, we
first replace \texttt{UNK} predictions
with their corresponding input words
and then evaluate the results.
We evaluate models for accuracy 
(percentage of correct words among all words)
and word correction rate
(percentage of misspelt tokens corrected).
We use 
AllenNLP\footnote{\href{https://allennlp.org/elmo}{allennlp.org/elmo}} 
and 
Huggingface\footnote{\href{https://huggingface.co/transformers/model\_doc/bert.html}{huggingface.co/transformers/model\_doc/bert.html}} 
libraries to use ELMo and BERT respectively.
All neural models in our toolkit are implemented using 
the Pytorch library \cite{paszke2017automatic},
and are compatible to run on both 
CPU and GPU environments.
Performance of different models
are presented in Table~\ref{tab:table-0-unnamed}.

\begin{table*}[!ht]
\small
\begin{tabular}{l|rr|rr|rr}
\multicolumn{7}{c}{\small{\textbf{Spelling correction systems in NeuSpell} (Word-Level Accuracy / Correction Rate)}} \\ \hline 
\multicolumn{1}{c|}{} & \multicolumn{2}{c}{Synthetic} & \multicolumn{2}{c}{Natural} & \multicolumn{2}{c}{Ambiguous} \\ 
\multicolumn{1}{c|}{} & \multicolumn{1}{c}{\textsc{word-test}} & \multicolumn{1}{c}{\textsc{prob-test}} & \multicolumn{1}{c}{\textsc{BEA-60K}} & \multicolumn{1}{c}{\textsc{JFLEG}} & \multicolumn{1}{c}{\textsc{BEA-4660}} & \multicolumn{1}{c}{\textsc{BEA-322}}  \\
\hline
\textsc{Aspell} \cite{aspell} 
& 43.6 / 16.9 & 47.4 / 27.5 & 68.0 / 48.7 & 73.1 / 55.6 & 68.5 / 10.1 & 61.1 / 18.9  \\ 
\textsc{JamSpell} \cite{jamspell} 
& 90.6 / 55.6 & 93.5 / 68.5 & 97.2 / 68.9 & 98.3 / 74.5 & \textbf{98.5} / 72.9 & \textbf{96.7} / 52.3 \\
\hline
\textsc{char-cnn-lstm} \cite{kim2015characteraware} 
& 97.0 / 88.0 & 96.5 / 84.1 &  96.2 / 75.8 & 97.6 / 80.1 & 97.5 / 82.7 & 94.5 / 57.3  \\ 
\textsc{sc-lstm} \cite{sakaguchi2016robsut} 
& 97.6 / 90.5 & 96.6 / 84.8 & 96.0 / 76.7 & 97.6 / 81.1 & 97.3 / 86.6 & 94.9 / 65.9  \\ 
\textsc{char-lstm-lstm} \cite{li2018spelling} 
& 98.0 / 91.1 & 97.1 / 86.6 & 96.5 / 77.3 & 97.6 / 81.6 & 97.8 / 84.0 & 95.4 / 63.2  \\ 
\textsc{Bert} \cite{devlin2018bert}
& \textbf{98.9} / \textbf{95.3} &  \textbf{98.2} / \textbf{91.5} & 93.4 / 79.1  & \textbf{97.9} / \textbf{85.0} & 98.4 / \textbf{92.5} & 96.0 / \textbf{72.1} \\ 
\hline
\textsc{sc-lstm} &    &   &  &  & &  \\ 
\hspace{3mm}+ \textsc{Elmo} (input) 
& 98.5 / 94.0 & 97.6 / 89.1 & 96.5 / \textbf{79.8} & 97.8 / \textbf{85.0} & 98.2 / 91.9 & 96.1 / 69.7  \\ 
\hspace{3mm}+ \textsc{Elmo} (output)  
& 97.9 / 91.4  &  97.0 / 86.1 & \textbf{98.0} / 78.5 & 96.4 / 76.7 & 97.9 / 88.1 & 95.2 / 63.2  \\ 
\hspace{3mm}+ \textsc{Bert} (input)
& 98.7 / 94.3  & 97.9 / 89.5 & 96.2 / 77.0 & 97.8 / 83.9 & 98.4 / 90.2 & 96.0 / 67.8  \\ 
\hspace{3mm}+ \textsc{Bert} (output)
& 98.1 / 92.3 &  97.2 / 86.9 & 95.9 / 76.0 &  97.6 / 81.0 & 97.8 / 88.1 & 95.1 / 67.2 \\ 
\hline
\end{tabular}
\caption{Performance of different models in NeuSpell on natural, synthetic, and ambiguous test sets. All models are trained using \textsc{Prob}+\textsc{Word} noising strategy.}
\label{tab:table-1-comparative-analysis}
\end{table*}
\section{Synthetic Training Datasets}
\label{sec:datasets}

Due to scarcity of available 
parallel data for
spelling correction, 
we noise sentences 
to generate misspelt-correct sentence pairs.
We use $1.6$M sentences from 
the one billion word benchmark
\citep{chelba2013billion} dataset 
as our clean corpus. 
Using different noising strategies from existing literature,
we noise 
$\sim$20\% of the 
tokens in the clean corpus by injecting spelling mistakes in each sentence. Below, we briefly describe these strategies.

\paragraph{\textsc{Random:}} Following 
\citet{sakaguchi2016robsut}, 
this noising strategy 
involves four character-level operations: permute, delete, 
insert and replace. We
manipulate 
only the internal characters 
of a word.
The permute operation
jumbles a pair of 
consecutive characters,  
delete operation 
randomly deletes 
one of the characters, 
insert operation
randomly inserts 
an alphabet and
replace operation
swaps a character 
with a randomly selected 
alphabet. 
For every word in the clean corpus,
we select one of the four operations 
with $0.1$ probability each. 
We do not modify
words of length three or smaller.

\paragraph{\textsc{Word}:}
Inspired 
from \citet{belinkov2017synthetic}, we 
swap a word with its 
noised counterpart from a pre-built
lookup table. We collect $109$K 
misspelt-correct word pairs for
$17$K popular English words 
from a variety of 
public 
sources.\footnote{\href{https://en.wikipedia.org/wiki/Wikipedia:Lists_of_common_misspellings/For_machines}{https://en.wikipedia.org/}, \href{https://www.dcs.bbk.ac.uk/~ROGER/corpora.html}{dcs.bbk.ac.uk}, \href{http://norvig.com/ngrams/spell-errors.txt}{norvig.com}, \href{https://corpus.mml.cam.ac.uk/efcamdat}{corpus.mml.cam.ac.uk/efcamdat}}

For every word in the clean corpus, 
we replace it 
by a 
random misspelling 
(with a probability of $0.3$)
sampled from all the 
misspellings associated 
with that word in the lookup table. Words not present in the lookup table are left as is.

\paragraph{\textsc{Prob}:} 
Recently, \citet{Piktus_2019} 
released a corpus of $20$M
correct-misspelt word pairs,
generated from 
logs of a search engine.\footnote{\label{moe_footnote}\href{https://github.com/facebookresearch/moe}{https://github.com/facebookresearch/moe}}
We use this corpus
to 
construct a character-level
confusion dictionary 
where the keys are 
$\langle$character, context$\rangle$ pairs 
and the values are a list 
of potential character replacements with their frequencies. 
This dictionary is subsequently used 
to sample character-level errors in a given context.
We use a 
context of $3$ characters, and backoff to $2, 1$, and $0$ characters.
Notably, due to the large number of 
unedited characters in the corpus, 
the most probable replacement
will often be the  
same as the source character.

\paragraph{\textsc{Prob+Word}:}
For this strategy, we simply concatenate the 
training data obtained from both 
\textsc{Word} and 
\textsc{Prob}
strategies.

\section{Evaluation Benchmarks}
\label{sec:evaluation_benchmarks}

\paragraph{Natural misspellings in context}
Many publicly 
available spell-checkers correctors
evaluate 
on isolated misspellings~\cite{aspell, rogers, norvig}.
Whereas, we evaluate our systems 
using misspellings in context, 
by using publicly available
datasets for the task of Grammatical 
Error Correction (GEC).
Since the GEC datasets 
are annotated 
for various types of grammatical mistakes, 
we only sample
errors of \texttt{SPELL} type.

Among the GEC datasets in
BEA-2019 shared task\footnote{\href{https://www.cl.cam.ac.uk/research/nl/bea2019st/}{www.cl.cam.ac.uk/research/nl/bea2019st/}},
the Write \& Improve (W\&I) dataset
along with the LOCNESS dataset
are a collection of texts in English (mainly essays)
written by language learners with varying proficiency levels
\citep{bryant-etal-2019-bea, Granger1998TheCL}.
The First Certificate in English (FCE)
dataset is another collection of essays
in English written by non-native learners 
taking a language assessment exam
\citep{yannakoudakis-etal-2011-new}
and
the Lang-8 dataset 
is a collection of English texts from Lang-8
online language learning website
\citep{mizumoto-etal-2011-mining, tajiri-etal-2012-tense}.
We combine data from these four sources 
to create the \textsc{BEA-60k}
test set with nearly
70K spelling mistakes (6.8\% of all tokens) 
in 63044 sentences.

The JHU 
FLuency-Extended GUG Corpus (\textsc{JFLEG})
dataset \citep{napoles-etal-2017-jfleg}
is another collection of essays written by  
English learners with different first languages.
This dataset contains 
$2$K spelling mistakes (6.1\% of all tokens)
in 1601 sentences.
We use the
\textsc{BEA-60k} and \textsc{JFLEG}
datasets only for 
the purposes of evaluation, 
and do not use them in
training process.

\paragraph{Synthetic misspellings in context}
From the two noising strategies
described in \S\ref{sec:datasets},
we additionally create two test
sets: \textsc{word-test} and \textsc{prob-test}.
Each of these test sets
contain around $1.2$M 
spelling mistakes ($19.5$\% of all tokens) in $273$K sentences.

\paragraph{Ambiguous misspellings in context}
Besides the natural and synthetic
test sets,
we create a challenge set 
of ambiguous spelling mistakes, \emph{which require additional context to unambiguously correct them}.
For instance, the word  
\uwave{whitch} can be corrected to ``witch'' or ``which'' 
depending upon the context.
Simliarly, for the word \uwave{begger}, both ``bigger'' or ``beggar'' can be appropriate corrections.
To create this challenge set, 
we 
select all such misspellings 
which are either $1$-edit distance away
from two (or more) legitimate dictionary words,
or have the same phonetic encoding 
as two (or more) dictionary words.
Using these two criteria, 
we sometimes end up with inflections 
of the same word, hence we use a stemmer and lemmatizer from the NLTK library
to weed those out.
Finally, we
manually prune
down the list to 322 sentences,
with one ambiguous mistake per sentence.
We refer to this set as \textsc{BEA-322}.

We also create another larger test set where we artificially misspell
two different words in sentences to their common ambiguous misspelling. This process 
results in a set with $4660$ misspellings in $4660$ sentences, and is thus referred as \textsc{BEA-4660}.
Notably, for both these ambiguous test sets, a spelling correction system 
that doesn't use any context information
can at best correct 50\% of the mistakes.

\section{Results and Discussion}
\label{sec:experiments}

\begin{table*}[ht]
\small
\centering
\resizebox{\textwidth}{!}{%
\begin{tabular}{lllllll}
\multicolumn{7}{c}{\small{\textbf{Sentiment Analysis} (1-char attack / 2-char attack)}} \\ \hline
\multicolumn{1}{c}{\textbf{\small{Defenses}}} & \multicolumn{1}{c}{\textbf{\small{No Attack}}} & \multicolumn{1}{c}{\textbf{\small{Swap}}} & \multicolumn{1}{c}{\textbf{\small{Drop}}} & \multicolumn{1}{c}{\textbf{\small{Add}}} & \multicolumn{1}{c}{\textbf{\small{Key}}} & \multicolumn{1}{c}{\textbf{\small{All}}} \\ \hline
\multicolumn{7}{c}{\small{Word-Level Models}} \\ \hline 
\textsc{sc-lstm} \citep{Pruthi_2019} & 79.3 & \textbf{78.6 / 78.5} & 69.1 / 65.3 & 65.0 / 59.2 & 69.6 / 65.6 & 63.2 / 52.4 \\
\textsc{sc-lstm}+\textsc{Elmo}(input) (F) & \textbf{79.6} & 77.9 / 77.2 & \textbf{72.2 / 69.2} & \textbf{65.5 / 62.0} & \textbf{71.1 / 68.3} & \textbf{64.0 / 58.0} \\ \hline
\multicolumn{7}{c}{\small{Char-Level Models}} \\ \hline 
\textsc{sc-lstm} \citep{Pruthi_2019} & 70.3 & 65.8 / 62.9 & 58.3 / 54.2 & \textbf{54.0 / 44.2} & \textbf{58.8} / 52.4 & \textbf{51.6} / 39.8 \\ 
\textsc{sc-lstm}+\textsc{Elmo}(input) (F) & \textbf{70.9} & \textbf{67.0 / 64.6} & \textbf{61.2 / 58.4} & 53.0 / 43.0 & 58.1 / \textbf{53.3} & 51.5 / \textbf{41.0} \\ \hline
\multicolumn{7}{c}{\small{Word+Char Models}} \\ \hline 
\textsc{sc-lstm} \citep{Pruthi_2019} & 80.1 & 79.0 / 78.7 & 69.5 / 65.7 & 64.0 / \textbf{59.0} & 66.0 / 62.0 & 61.5 / \textbf{56.5} \\
\textsc{sc-lstm}+\textsc{Elmo}(input) (F) & \textbf{80.6} & \textbf{79.4 / 78.8} & \textbf{73.1 / 69.8} & \textbf{66.0} / 58.0 & \textbf{72.2 / 68.7} & \textbf{64.0} / 54.5 \\ \hline
\end{tabular}%
}
\caption{We evaluate spelling correction systems in NeuSpell against adversarial misspellings.}
\label{tab:table-3-adversarial}
\end{table*}

\subsection{Spelling Correction }
\label{ssec:exp_spell_correction}

We evaluate the $10$ spelling correction systems in NeuSpell across $6$ different datasets (see Table~\ref{tab:table-1-comparative-analysis}).
Among the spelling correction systems, all the neural models in the toolkit
are trained using synthetic training dataset,
using the
\textsc{Prob}+\textsc{Word} synthetic data.
We use the recommended configurations for
Aspell and Jamspell, but do not fine-tune them on our synthetic dataset.
In all our experiments, 
vocabulary of neural models
is restricted to the top 100K 
frequent words of the clean corpus.

We observe that although off-the-shelf 
checker Jamspell leverages context,
it is often inadequate.
We see that models
comprising of deep contextual
representations
consistently outperform other 
existing neural models for 
the spelling correction task.
We also note that the \textsc{BERT} model performs consistently well 
across all our benchmarks. 
For the ambiguous \textsc{BEA-322} test set,
we manually evaluated corrections from Grammarly---a professional 
paid service for assistive writing.%
\footnote{Retrieved on July 13, 2020
.}
We found that our best model for this set, i.e. \textsc{BERT}, outperforms corrections from Grammarly ($72.1$\% vs $71.4$\%) 
We attribute 
the success of 
our toolkit's well
performing models to
(i) better 
representations of the context, from large pre-trained
models; 
(ii) swap invariant semi-character 
representations; and
(iii) training models with synthetic data
consisting of noise patterns from real-world
misspellings.
We follow up these results
with an ablation study to understand the 
role
of each noising strategy
(Table~\ref{tab:table-2-training-strategies}).\footnote{To fairly compare across
different noise types, in this experiment
we include only 50\% of samples from each of \textsc{Prob}
and \textsc{Word} noises to construct the \textsc{Prob}+\textsc{Word}
noise set.}
For each of the $5$ models
evaluated, we observe that 
models trained with
\textsc{Prob} noise outperform
those trained with \textsc{Word} or \textsc{Random} noises.
Across all the models, we further observe that using \textsc{Prob}+\textsc{Word} strategy
improves correction rates by at least 10\% 
in comparison to \textsc{Random} noising.

\begin{table}[!htbp]
\resizebox{0.48\textwidth}{!}{%
    \begin{tabular}{llcc}
    \multicolumn{4}{c}{\textbf{Spelling Correction} (Word-Level Accuracy / Correction Rate)} \\ 
    \hline
    \multirow{2}{*}{Model} & \multirow{1}{*}{Train Noise} & \multicolumn{2}{c}{Natural test sets} \\
    &  & \multicolumn{1}{c}{\textsc{BEA-60k}} & \multicolumn{1}{c}{JFLEG} \\ 
    \hline
    \multirow{2}{*}{\makecell[l]{\textsc{char-cnn-lstm}\\\cite{kim2015characteraware}}} & \textsc{Random} & 95.9 / 66.6 & 97.4 / 69.3 \\
    & \textsc{Word} & 95.9 / 70.2 & 97.4 / 74.5 \\ 
    & \textsc{Prob} & 96.1 / 71.4 & 97.4 / 77.3 \\ 
    & \textsc{Prob}+\textsc{Word} & 96.2 / 75.5 & 97.4 / 79.2 \\
    \hline
    \multirow{2}{*}{\makecell[l]{\textsc{sc-lstm}\\\cite{sakaguchi2016robsut}}} & \textsc{Random} & 96.1 / 64.2 & 97.4 / 66.2 \\ 
    & \textsc{Word} & 95.4 / 68.3 & 97.4 / 73.7 \\ 
    & \textsc{Prob} & 95.7 / 71.9 & 97.2 / 75.9 \\ 
    & \textsc{Prob}+\textsc{Word} & 95.9 / 76.0 & 97.6 / 80.3 \\
    \hline
    \multirow{2}{*}{\makecell[l]{\textsc{char-lstm-lstm}\\\cite{li2018spelling}}} & \textsc{Random} & 96.2 / 67.1 & 97.6 / 70.2\\ 
    & \textsc{Word} & 96.0 / 69.8 & 97.5 / 74.6 \\ 
    & \textsc{Prob} & 96.3 / 73.5 & 97.4 / 78.2 \\ 
    & \textsc{Prob}+\textsc{Word} & 96.3 / 76.4 & 97.5 / 80.2 \\
    \hline
    \multirow{2}{*}{\makecell[l]{\textsc{bert}\\\cite{devlin2018bert}}} & \textsc{Random} & \textbf{96.9} / 66.3 & \textbf{98.2} / 74.4 \\ 
    & \textsc{Word} & 95.3 / 61.1 & 97.3 / 70.4 \\ 
    & \textsc{Prob} & 96.2 / 73.8 & 97.8/ 80.5 \\ 
    & \textsc{Prob}+\textsc{Word}  & 96.1 / 77.1 & 97.8 / 82.4 \\
    \hline
    \multirow{1}{*}{\textsc{sc-lstm}} & \textsc{Random} & \textbf{96.9} / 69.1 & 97.8 / 73.3 \\ 
    \hspace{3mm}+ \textsc{Elmo} (input) & \textsc{Word} & 96.0 / 70.5 & 97.5 / 75.6 \\ 
    & \textsc{Prob} & 96.8 / 77.0 & 97.7 / 80.9 \\ 
    & \textsc{Prob}+ \textsc{Word}  & 96.5 / \textbf{79.2} & 97.8 / \textbf{83.2} \\ \hline
    \end{tabular}
}
\caption{Evaluation of models on the natural test sets when trained using synthetic datasets curated using different noising strategies.}
\label{tab:table-2-training-strategies}
\end{table}

\subsection{Defense against Adversarial Mispellings}
\label{ssec:exp_defense}

Many recent studies 
have demonstrated 
the susceptibility of 
neural models under 
word- and character-level attacks 
~\cite{alzantot-etal-2018-generating, belinkov2017synthetic, Piktus_2019, Pruthi_2019}.
To combat adversarial misspellings, \citet{Pruthi_2019}
find spell checkers to be a viable defense. Therefore, 
we also evaluate spell checkers in our toolkit against
adversarial misspellings.

We follow the same experimental setup as \citet{Pruthi_2019} for the sentiment classification task under different adversarial attacks. We finetune \textsc{sc-lstm}+\textsc{Elmo}(input) model on movie reviews data from the Stanford Sentiment Treebank (SST) \citep{socher-etal-2013-recursive}, using the same noising strategy as in~\cite{Pruthi_2019}. 
As we observe from Table~\ref{tab:table-3-adversarial}, our corrector from NeuSpell toolkit (\textsc{sc-lstm}+\textsc{Elmo}(input)(F)) outperforms the spelling corrections models proposed in \cite{Pruthi_2019} in most cases. 

\section{Conclusion}
\label{sec:conclusion}
In this paper, we describe 
NeuSpell, a spelling correction 
toolkit, comprising
ten different
models.
Unlike popular open-source spell checkers, our models 
accurately capture the context around the misspelt words.
We also supplement models in our toolkit with a unified 
command line, and a web interface.
The toolkit is open-sourced, free for public use, and available at \url{https://github.com/neuspell/neuspell}. A demo of the trained spelling correction models  can be accessed at
\url{https://neuspell.github.io/}.

\section*{Acknowledgements}
The authors thank Punit Singh Koura for insightful discussions and participation during the initial phase of the project.

\bibliography{emnlp2020}
\bibliographystyle{acl_natbib}

\end{document}